\documentclass[conference]{IEEEtran}
\IEEEoverridecommandlockouts
\usepackage{cite}
\usepackage{amsmath,amssymb,amsfonts}
\usepackage{algorithmic}
\usepackage{graphicx}
\usepackage{textcomp}
\usepackage{xcolor}
\usepackage{booktabs}
\usepackage[pagebackref,breaklinks,colorlinks]{hyperref}

\def\BibTeX{{\rm B\kern-.05em{\sc i\kern-.025em b}\kern-.08em
    T\kern-.1667em\lower.7ex\hbox{E}\kern-.125emX}}
\begin{document}

\title{Combining Contrastive and Supervised Learning for~Video Super-Resolution Detection \\
}

\author{Viacheslav Meshchaninov\quad Ivan Molodetskikh\quad 
Dmitriy Vatolin \\
Lomonosov Moscow State University\\
Moscow\\
{\tt\small \{vyacheslav.meshchaninov, ivan.molodetskikh, dmitriy\}@graphics.cs.msu.ru}
}

\maketitle

\begin{abstract}
Upscaled video detection is a helpful tool in multimedia forensics, but it’s a challenging task that involves various upscaling and compression algorithms. 
There are many resolution-enhancement methods, including interpolation and deep-learning-based super-resolution, and they leave unique traces. 
In this work, we propose a new upscaled-resolution-detection method based on learning of visual representations using contrastive and cross-entropy losses. 
To explain how the method detects videos, we systematically review the major components of our framework—in particular, we show that most data-augmentation approaches hinder the learning of the method. 
Through extensive experiments on various datasets, we demonstrate that our method effectively detects upscaling even in compressed videos and outperforms the state-of-the-art alternatives.
The code and models are publicly available at \url{https://github.com/msu-video-group/SRDM} 
\end{abstract}

\begin{IEEEkeywords}
native resolution, upscaling detection, super-resolution, interpolation, video compression 
\end{IEEEkeywords}

\section{Introduction}
Recent years have seen a big jump in the quality of super-resolution, and these models continue to improve.
Architectures ranging from transformers~\cite{liang2021swinir,parmar2018image} to recurrent neural networks~\cite{haris2019recurrent,chadha2020iseebetter,isobe2020video} and generative adversarial networks~\cite{pan2021deep, su2020local, wang2018esrgan, tian2020tdan, wang2021real} have been proposed. 
These methods can generate high-quality videos and images that the human eye can barely distinguish.
Malefactors, such as dishonest camera sellers and unscrupulous videographers, can take advantage of such capabilities, leading to ethical and legal problems~\cite{su2017hierarchical, zhang2020robustness}.
Therefore, blind detection of video upscaling is crucial.

Prior works\cite{yang2020native, zhang2020robustness, su2017hierarchical, cao2019resampling} mainly focus on detection of classic interpolation algorithms, such as nearest-neighbor interpolation, bilinear interpolation, and bicubic interpolation~\cite{keys1981cubic}. 
Yang et al.~\cite{yang2020native} targeted super-resolution upscalers, but their test set contains only 10 videos and three super-resolution methods, limiting their ability to generalize. 
Also, they neglected to consider a popular practical case: compressed videos.

Like deepfake detection, super-resolution detection must be able to discern generated data.
In practice, however, methods designed for one task may fail to work on others. 
One recent deepfake-detection method \cite{wang2020cnn}, for example, tried to identify state-of-the-art super-resolution, but it has only achieved a mediocre 93.6\% average precision for original images and 78.1\% for JPEG-compressed images.

This work presents an approach to detecting both compressed and uncompressed upscaled videos.
It incorporates supervised and contrastive learning \cite{hadsell2006dimensionality, dosovitskiy2014discriminative, chen2020simple, bardes2021vicreg}, which increases the accuracy of the results predominantly for compressed videos, as we show further.
We trained our classifier by generating numerous upscaled and compressed videos using several super-resolution architectures.
To evaluate it, we chose the test portion of the REDS dataset~\cite{nah2019ntire} and 100 videos from the Vimeo-90K\cite{xue2019video}, as well as six super-resolution models with capabilities ranging from generation of a single image to generation of multiple continuous video frames. 
These models are LGFN\cite{su2020local}, RBPN\cite{haris2019recurrent}, Real-ESRGAN\cite{wang2021real}, RRN\cite{isobe2020revisiting}, SOF-VSR\cite{wang2020deep}, and Topaz\cite{topaz}. 
We evaluated our algorithm on the MSU Video Super-Resolution~\cite{lyapustin2022towards} and RealSR~\cite{cai2019toward} benchmarks; it achieved good accuracy on this new data, detecting 30 out of 32 upscaling methods and thereby confirming its ability to generalize.

In summary, our main contribution is a new method for detecting upscaled videos that combines the ideas of contrastive and supervised learning. 
We also introduced the strategy and details of the data-processing pipeline.

\section{Proposed Method}
In this section, we first describe our proposed method’s overall architecture. 
Next, we introduce the design motivation for the loss function and detail the learning process.

\subsection{Overall Architecture}

\begin{figure*}[t]
    \begin{center}
        \includegraphics[width=\linewidth]{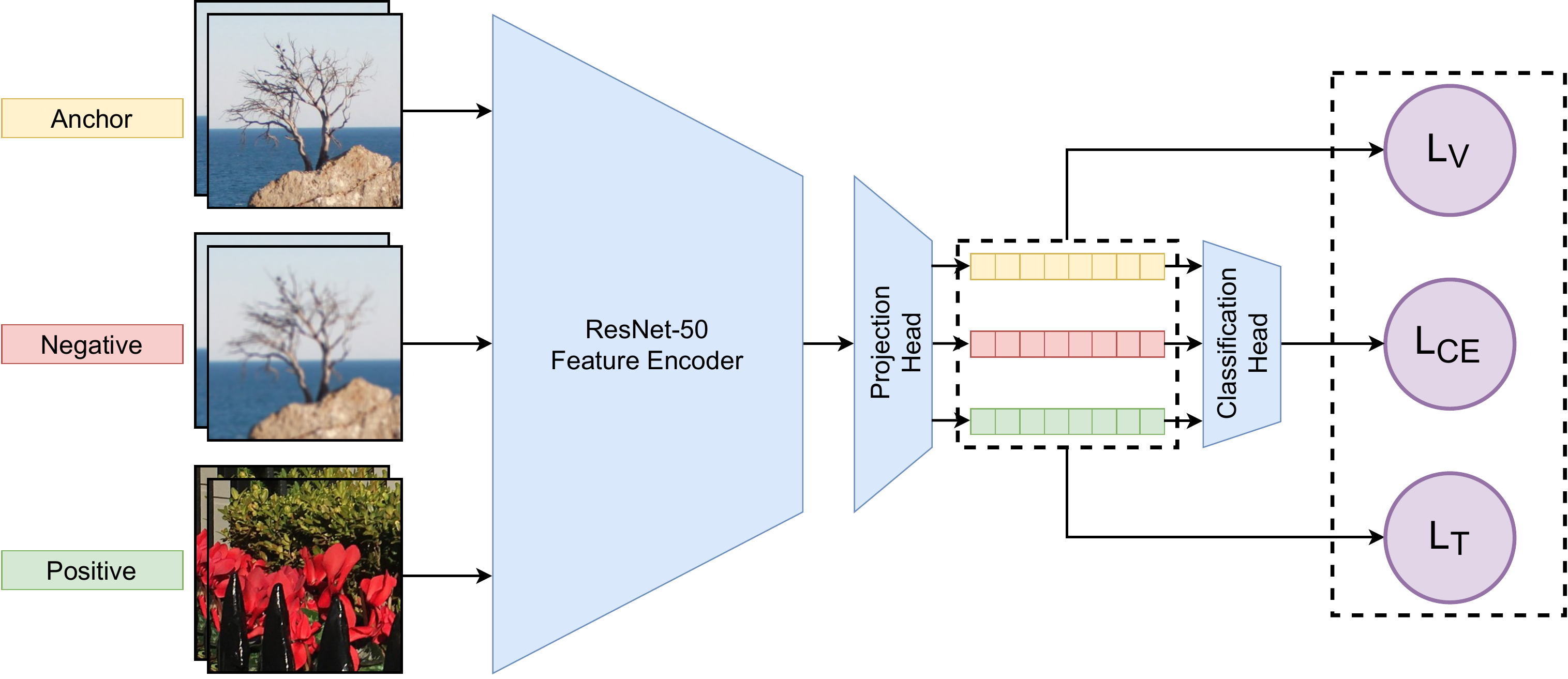}
    \end{center}
    \vspace*{-1\baselineskip}
    \caption{Architecture of the proposed method}%
    \label{fig:overall_architecture}
\end{figure*}

Inspired by recent contrastive-learning algorithms\cite{hadsell2006dimensionality, dosovitskiy2014discriminative, chen2020simple, bardes2021vicreg}, we propose an architecture consisting of four major components: a data-preprocessing pipeline, a feature encoder, a projection head, and a classification head. Figure~\ref{fig:overall_architecture} shows an overview of the architecture.

\begin{itemize}
\item
The data-preprocessing pipeline includes a data-augmentation module that takes as input two consecutive video frames.
It then sequentially applies two simple augmentations: The first is random cropping with a size of 224 and cutout~\cite{devries2017improved}.
As we show in Section 3, the combination of a random cutout and video compression is crucial to achieving good performance.
The addition of other augmentations, such as JPEG compression, Gaussian noise, and Gaussian blur, only decreases the model performance.

\item
The neural-network-based feature encoder $f(x)$ extracts representation vectors from augmented-data samples.
We use ResNet-50\cite{he2016deep} to obtain the representation $h = f(x)$, where $x$ is the augmented-data example and $h$ is the output of the average-pooling layer.
To accommodate two frames, we increase the first layer’s channel count from three to six.

\item
The projection head $g(h)$ is a small neural network that maps representations to the space in which contrastive loss is applied. 
We use an MLP with three hidden layers to obtain the projection $z = g(h)$.
This component is inspired by SimCLR~\cite{chen2020simple}, which showed that defining contrastive loss on $z$ rather than $h$ is beneficial.

\item
The classification head $c(z)$ is also a small neural network with a similar architecture.
It maps projection $z$ to the probability that the frames are upscaled.
\end{itemize}

\subsection{Loss Function}
Model optimizes the sum of three losses: cross-entropy, triplet, and variance.
We randomly sampled from a video with real resolution a pair of consecutive frames, which we consider to be the anchor sample $a$, then we took their upscaled version, which we consider to be the negative sample $n$.
Finally, we randomly chose from another video with real resolution a pair of consecutive frames and consider them to be the positive sample $p$. 
After receiving their representation from the encoder: $f(a)$, $f(n)$, and $f(p)$ -- the projector maps them to the space in which contrastive loss is applied.

Eventually we obtain $h_{a} = g(f(a))$, $h_{n} = g(f(n))$, and $h_{p} = g(f(p))$. 
Let $sim(u, v) = \frac{u^T v}{||u|| ||v||}$ denote the dot product of two normalized $u$ and $v$ vectors (i.e., their cosine similarity). 
Then the triplet loss is the following:
$$L_{T} = \frac{1}{N} \sum^{N}_{i=1} (sim(h_{a_i}, h_{p_i}) - sim(h_{a_i}, h_{n_i}) + m),$$
where $N$ is the batch size and $m$ is the margin.

This criterion tends to make the representation of the anchor video closer in cosine similarity to the representation of the randomly sampled video, which has completely different content.
And it tends to make the representation of the anchor video differ more from the representation of the upscaled video, which has the same content but underwent processing by an upscaler.

We denote $H_a = [h_{a_1}, ..., h_{a_N}]$, $H_n = [h_{n_1}, ..., h_{n_N}]$, and $H_p = [h_{p_1}, ..., h_{p_N}]$ as the three batches of $N$ vectors of dimension $d$. 
We also denote by $h^j$ the vector comprising each value at dimension $j$ in all vectors in $H \in \{H_a, H_n, H_p\}$. 
As in \cite{bardes2021vicreg}, we define the variance-regularization term $v$ as a hinge function on the standard deviation of the embeddings along the batch dimension:
$$v(H) = \frac{1}{d} \sum^{d}_{j=1} \max(0, \gamma - S(h^j, \varepsilon)),$$
where $S$ is the the regularized standard deviation defined by
$$S(x, \varepsilon) = \sqrt{Var(x) + \varepsilon}.$$
Here, $\gamma$ is a constant target value for the standard deviation, fixed to 1 in our experiments, and $\varepsilon$ is a small scalar that prevents numerical instabilities. 
We then define the variance loss as follows:
$$L_{V} = v(H_a) + v(H_n) + v(H_p)$$
This criterion encourages the variance in the current batch to equal $\gamma$ along each dimension, preventing a collapse in which all the inputs map to the same vector. 

We denote $C_a = [c{h_{a_1}}, ..., c{h_{a_N}}]$, $C_n = [c{h_{n_1}}, ..., c{h_{n_N}}]$, and $C_p = [c{h_{p_1}}, ..., c{h_{p_N}}]$ as the three batches of logits applying the classification head; the cross-entropy loss is therefore
$$L_{CE} = \frac{1}{4}CE(C_a, 0) + \frac{1}{4}CE(C_p, 0) + \frac{1}{2}CE(C_n, 1),$$
where $CE(y, t)$ is the cross-entropy between the output $y$ and target $t$.
We use coefficients $\frac{1}{4}$, $\frac{1}{4}$, and $\frac{1}{2}$ for the anchors, positives, and negatives, respectively, to eliminate bias between videos of real resolution and those of fake resolution.

The overall loss function is the sum of the cross-entropy, triplet, and variance losses:
$$Loss = L_{CE} + L_{T} + L_{V}.$$

\subsection{Train Dataset}

To train our networks, we used the REDS \cite{nah2019ntire} dataset, which consists of 240 original video sequences, each containing 500 frames.
We divided the sequences into five blocks of 100 frames each and took 20 frames, the 10th to the 29th, inclusive, from each block.
Then we used the x264 video encoder to compress the original videos with a uniformly random CRF between 15 and 30, inclusive.
These sequences constitute the “real-resolution” dataset. 
The next step was to downscale all the original sequences using bilinear interpolation and upscale them using six super-resolution methods: Real-ESRGAN\cite{wang2021real}, ESRGAN\cite{wang2018esrgan}, RRN\cite{isobe2020revisiting}, RBPN\cite{haris2019recurrent}, SOF-VSR\cite{wang2020deep}, and RealSR\cite{ji2020real}. 
We compressed the upscaled videos in the same way as the originals. 
The upscaled videos constitute the “fake-resolution” dataset.

\section{Experiments}

In this section, we first explore the optimal settings for our proposed super-resolution-detection method and then present experimental results to demonstrate its effectiveness.

\subsection{Implementation Details}
To create the input for our feature extractor, we stack two consecutive video frames.
Our method uses a pretrained ResNet-50 network as a backbone, engages in a warm start using ImageNet\cite{deng2009imagenet} pre-trained weights, employs a three-layer MLP projector to project the representation to 128-dimensional latent space, and includes a three-layer MLP classification head.
We apply the AdamW \cite{loshchilov2017decoupled} optimizer for 300 epochs at a learning rate of $2 \times 10^{-5}$ and weight decay of $0.05$. 
The batch size is 64, the algorithm implements a cosine-decay learning rate with a 20-epoch linear warmup and the initial learning rate is $5 \times 10^{-6}$. 

\subsection{Evaluation Metrics}

We chose the standardized balanced accuracy (b-Accuracy) and F1-score to estimate our models. 
These metrics require that detectors maintain a tiny false-positive rate, especially in the practical case of automatically screening for fakes on social media.
Also, to test our method separately on each super-resolution method, we used accuracy.

\subsection{Test Dataset}

\begin{table}[htbp]
\caption{Fake-resolution recognition accuracy}
\begin{center}
\begin{tabular}{l|c|c|c|c}
  \toprule
    SR method & \multicolumn{2}{|c|}{No compression} & \multicolumn{2}{|c}{Compression} \\
    \cline{2-5}
    & Accuracy & F1-score & Accuracy & F1-score  \\
  \midrule
    \multicolumn{5}{c}{Our method}\\
    \cline{1-5}
    Topaz & 0.995 & 0.959 & 0.991 & 0.972\\
    LGFN & 0.943 & 0.934 & 0.895 & 0.886 \\
    RBPN & 0.957 & 0.941 & 0.904 & 0.895\\
    Real-ESRGAN & 1.000 & 0.963 & 0.967 & 0.956\\
    RRN & 0.977 & 0.951 & 0.938 & 0.945\\
    SOF-VSR-BD & 0.981 & 0.910 & 0.909 & 0.930\\
    \cline{1-5}
    \multicolumn{5}{c}{Cao et al. \cite{cao2019resampling}}\\
    \cline{1-5}
    Topaz & 0.828 & 0.864 & 0.824 & 0.856 \\
    LGFN & 0.871 & 0.888 & 0.838 & 0.865 \\
    RBPN & 0.880 & 0.893 & 0.842 & 0.867 \\
    Real-ESRGAN & 0.788 & 0.833 & 0.927 & 0.862 \\
    RRN & 0.914 & 0.919 & 0.874 & 0.884 \\
    SOF-VSR-BD & 0.900 & 0.904 & 0.871 & 0.884 \\
    
  \bottomrule
  \end{tabular}
\label{tab:tab1}
\end{center}
\end{table}

Our tests took into account two video datasets: REDS and Vimeo-90K. 
REDS contains the realistic and dynamic scenes.
The Vimeo-90K is a large-scale high-quality video dataset.
The tests consisted of 30 sequences of 500 frames apiece.
We took the testing portion of the REDS dataset and 100 videos from Vimeo-90K and applied the same data-generation scheme we used for the training dataset, including compression. 
Our next step was to upscale the videos through several super-resolution methods: LGFN\cite{su2020local}, RBPN\cite{haris2019recurrent}, Real-ESRGAN\cite{wang2021real}, RRN\cite{isobe2020revisiting}, SOF-VSR\cite{wang2020deep}, and Topaz Gigapixel AI\cite{topaz}.
The results appear in Table~\ref{tab:tab1}.

\subsection{Evaluation Using MSU Video Super-Resolution Benchmark}

To evaluate the cross-dataset generalizability of our approach, we used the MSU Video Super-Resolution Benchmark~\cite{lyapustin2022towards} dataset, which contains the most complex content for the restoration task: faces, text, QR codes, car license plates, unpatterned textures, and small details.
Videos include different types of motion and different types of degradation.
The dataset contains 10 original real-resolution videos (GT), each corresponding to 32 videos upscaled using 32 different super-resolution methods.
The video is considered as fake-resolution video if method detects at least 5\% of all frames.
We tested our method on all the videos and calculated for each one the percent of detected super-resolution methods.
Our testing shows that the method detects real videos with 100\% accuracy. 
Table~\ref{tab:tab2} lists the comparison results for upscaled videos.

\begin{table}[htbp]
\caption{Evaluation using MSU VSR benchmark}
\begin{center}
\begin{tabular}{l|c|c}
  \toprule
    Part of the dataset scene & \multicolumn{2}{|c}{The percentage of detected SR methods} \\
    \cline{2-3}
    & Our method & Cao et al. \cite{cao2019resampling} \\
  \midrule
    GT & 1.000 & 0.375\\
    Board & 0.688 & 0.844\\
    QR & 0.688 & 0.906\\
    Text & 0.969 & 0.688 \\
    Tin foil & 0.938 & 0.844\\
    Color lines & 0.500 & 0.844\\
    License-plate numbers & 0.906 & 0.875\\
    Noise & 0.875 & 0.688 \\
    Resolution test chart & 0.906 & 0.906\\
  \bottomrule
  \end{tabular}
\label{tab:tab2}
\end{center}
\end{table}

\begin{table}[htbp]
\caption{Evaluation using RealSR benchmark}
\begin{center}
\begin{tabular}{l|c|c}
  \toprule
    Method & \multicolumn{2}{|c}{Accuracy} \\
    \cline{2-3}
    & Our method & Cao et al. \cite{cao2019resampling} \\
  \midrule
    GT & 0.877 & 0.734 \\
    RealSR & 0.993 & 0.938 \\
    Real-ESRGAN & 0.952 & 0.708 \\
    ESRGAN & 0.996 & 0.958\\
  \bottomrule
  \end{tabular}
\label{tab:tab3}
\end{center}
\end{table}

\begin{table}[htbp]
\caption{Effect of data augmentation}
\begin{center}
\begin{tabular}{l|c|c|c|c}
  \toprule
    Augmentation & \multicolumn{2}{|c|}{No compression} &  \multicolumn{2}{|c}{Compression} \\
    \cline{2-5}
    & b-Accuracy & F1-score & b-Accuracy & F1-score  \\
  \midrule
    No augmentation & 0.924 & 0.925 & 0.874 & 0.883 \\
    Blur & 0.923 & 0.937 & 0.879 & 0.885 \\
    JPEG & 0.914 & 0.929 & 0.867 & 0.878 \\
    Gaussian Noise & 0.918 & 0.937 & 0.875 & 0.885 \\
    Cutout & 0.951 & 0.952 & 0.933 & 0.933 \\
  \bottomrule
  \end{tabular}
\label{tab:tab4}
\end{center}
\end{table}

\subsection{Evaluation Using RealSR Benchmark}

We also used the RealSR benchmark to evaluate our method. 
We trained our method for image detection, changing only the input. 
Instead of two consecutive frames, the model receives just one frame.
The balance of the learning scheme remained the same.

We generated upscaled images using three state-of-the-art super-resolution techniques: RealSR \cite{cai2019toward}, Real-ESRGAN\cite{wang2021real}, and ESRGAN\cite{wang2018esrgan}.
Our estimate of the method’s quality on original (GT) and upscaled images is based on accuracy.
The results appear in Table~\ref{tab:tab3}.

\subsection{Effect of Data Augmentation}

Table~\ref{tab:tab4} quantifies the generalization ability of training through different augmentation methods.
We found that nearly all data-augmentation approaches hinder training. 
We believe blur, JPEG, noise, and other techniques change super-resolution traces and create new ones.
Also, we found that cutout  greatly increases our method’s performance on both compressed and uncompressed videos.

\subsection{Ablation Study}

To confirm the effectiveness of triplet and variance losses, we evaluated how they influence our model’s accuracy.
We compared regular ResNet-50 learning with cross-entropy loss (ResNet) and with the sum of the cross-entropy, triplet (ResNet-CT), and variance losses (ResNet-CTV). 
If a lightweight model was necessary—for example, in the practical case of automatic screening -- we trained both MobileNetV2~\cite{sandler2018mobilenetv2} and MobileNet-CTV.
We also compared our method with that of Cao et al. \cite{cao2019resampling}, which those we trained on the same data as our models.
The accuracy and F1-score results are in Table~\ref{tab:tab5}.

We also evaluated how the number of consecutive frames passed as the input influence our model's performance.
Table~\ref{tab:tab6} lists the comparison results.

\begin{table}[htbp]
\caption{Method evaluation using validation dataset}
\begin{center}
\begin{tabular}{l|c|c|c|c}
  \toprule
    Method & \multicolumn{2}{|c|}{No compression} &  \multicolumn{2}{|c}{Compression} \\
    \cline{2-5}
    & b-Accuracy & F1-score & b-Accuracy & F1-score  \\
  \midrule
    Cao et al. \cite{cao2019resampling} & 0.884 & 0.881 & 0.867 & 0.863\\
    MobileNetV2 & 0.900 & 0.908 & 0.894 & 0.901\\
    MobileNet-CTV & 0.929 & 0.917 & 0.920& 0.916\\
    ResNet & 0.930 & 0.936 & 0.915 & 0.912\\
    ResNet-CT & 0.935 & 0.937 & 0.918& 0.920\\
    ResNet-CTV & 0.948 & 0.948 & 0.925 & 0.924\\
  \bottomrule
  \end{tabular}
\label{tab:tab5}
\end{center}
\end{table}

\begin{table}[htbp]
\caption{The comparison of using different number of frames}
\begin{center}
\begin{tabular}{l|c|c|c|c}
  \toprule
    Number of frames & \multicolumn{2}{|c|}{No compression} &  \multicolumn{2}{|c}{Compression} \\
    \cline{2-5}
    & b-Accuracy & F1-score & b-Accuracy & F1-score  \\
  \midrule
    1 & 0.925 & 0.929 & 0.918 & 0.920 \\
    2 & 0.948 & 0.948 & 0.925 & 0.924 \\
    3 & 0.951 & 0.952 & 0.919 & 0.915 \\
  \bottomrule
  \end{tabular}
\label{tab:tab6}
\end{center}
\end{table}

\section{Conclusion}
In this paper, we proposed a novel approach to video-super-resolution detection that combines contrasting and supervised learning. 
The method first uses a ResNet backbone to extract deep features and then uses a lightweight MLP networks to project and classify representation of input frames. 
Our method achieved good performance in extensive experiments.

\section*{Acknowledgment}

This study was supported by Russian Science Foundation under grant 22-21-00478, https://rscf.ru/en/project/22-21-00478/.

Model training has been conducted on the high-performance IBM Polus cluster of the CS MSU faculty, http://hpc.cmc.msu.ru/polus.

{\small
\bibliographystyle{ieee_fullname}
\bibliography{main}
}

\end{document}